\documentclass[10pt, conference, compsocconf]{IEEEtran}
\usepackage{times}
\usepackage{latexsym}

\usepackage{url}

\usepackage{graphicx}
\usepackage{subcaption}
\usepackage{latexsym}
\usepackage{amsmath}
\usepackage{multirow}
\usepackage{tablefootnote}
\usepackage{comment}
\usepackage{amssymb}
\usepackage{url}
\let\labelindent\relax
\usepackage{enumitem}
\setlist{nosep}

\usepackage{graphicx,calc}
\newlength\myheight
\newlength\mydepth
\settototalheight\myheight{Xygp}
\newcommand*\inlinegraphics[1]{%
  \settototalheight\myheight{Xygp}%
  \settodepth\mydepth{Xygp}%
  \raisebox{-\mydepth}{\includegraphics[height=\myheight]{#1}}%
}

\DeclareMathOperator*{\argmax}{arg\,max}
\usepackage{enumitem}
\newcommand{\subscript}[2]{$#1 _ #2$}

\usepackage{cite}

\ifCLASSINFOpdf
\else
\fi
\begin{document}
%
\title{Decipherment of Historical Manuscript Images}



\author{\IEEEauthorblockN{Xusen Yin \& Nada Aldarrab}
\IEEEauthorblockA{Information Sciences Institute\\
University of Southern California\\
Marina del Rey, California, USA\\
\{xusenyin,aldarrab\}@isi.edu}

\and
\IEEEauthorblockN{Be\'ata Megyesi}
\IEEEauthorblockA{Department of Linguistics and Philology\\
Uppsala University\\
Uppsala, Sweden\\
Beata.Megyesi@lingfil.uu.se}

\and
\IEEEauthorblockN{Kevin Knight}
\IEEEauthorblockA{DiDi Labs\\
\\
Marina del Rey, California, USA\\
kevinknight@didiglobal.com}
}


%


\maketitle

\begin{abstract}
European libraries and archives are filled with enciphered manuscripts from the early modern period.  These include military and diplomatic correspondence, records of secret societies, private letters, and so on.  Although they are enciphered with classical cryptographic algorithms, their contents are unavailable to working historians.  We therefore attack the problem of automatically converting cipher manuscript images into plaintext.  We develop  unsupervised models for character segmentation, character-image clustering, and decipherment of cluster sequences. We experiment with both pipelined and joint models, and we give empirical results for multiple ciphers.
\end{abstract}


%
\IEEEpeerreviewmaketitle

\begin{figure*}
        \centering
        \begin{subfigure}[b]{0.5\textwidth}
            \centering
            \includegraphics[width=0.8\textwidth]{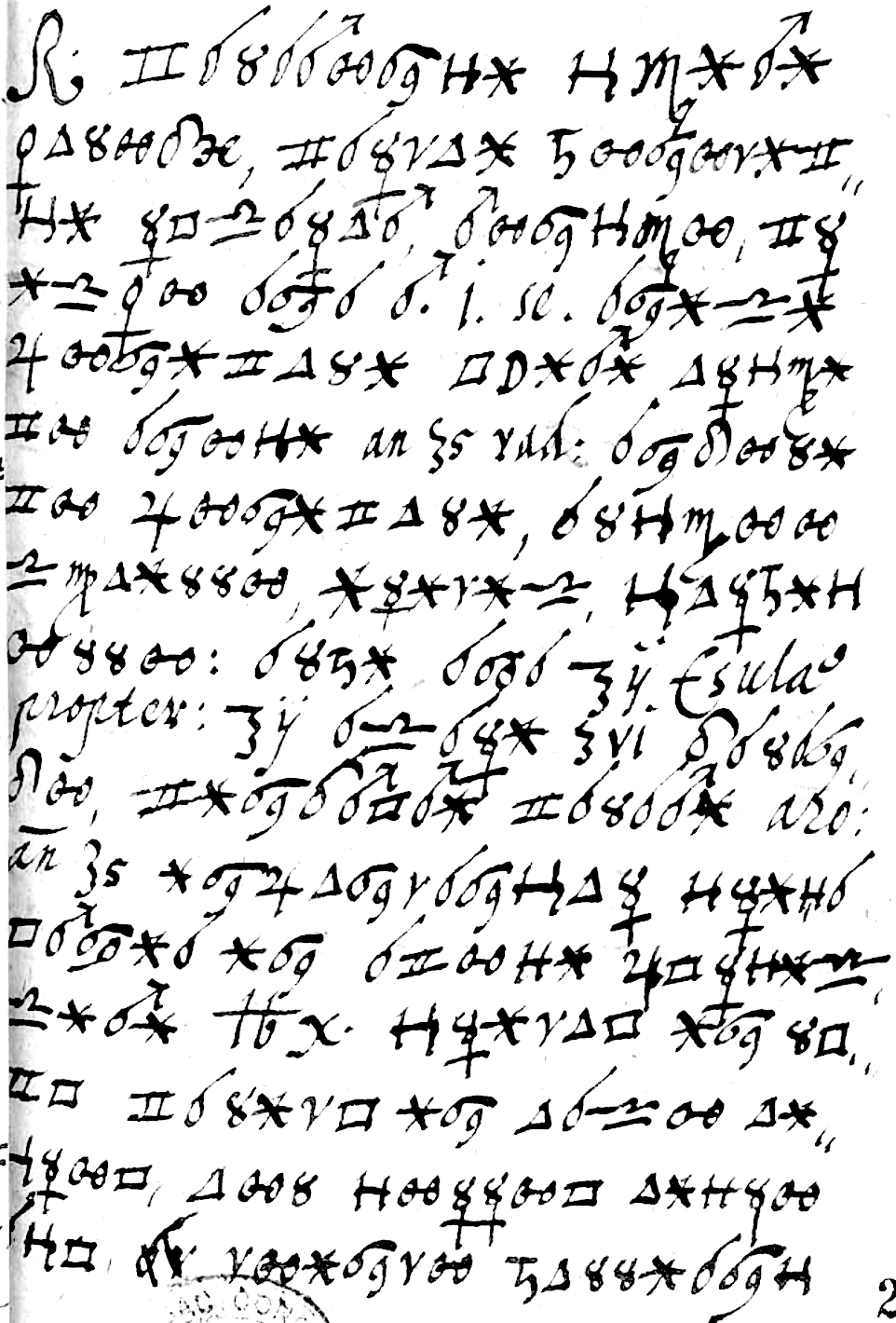}
            \caption[]{One page from Borg cipher}    
            \label{fig:one-page-borg}
        \end{subfigure}%
        \begin{subfigure}[b]{0.5\textwidth}  
            \centering 
            \includegraphics[width=0.8\textwidth]{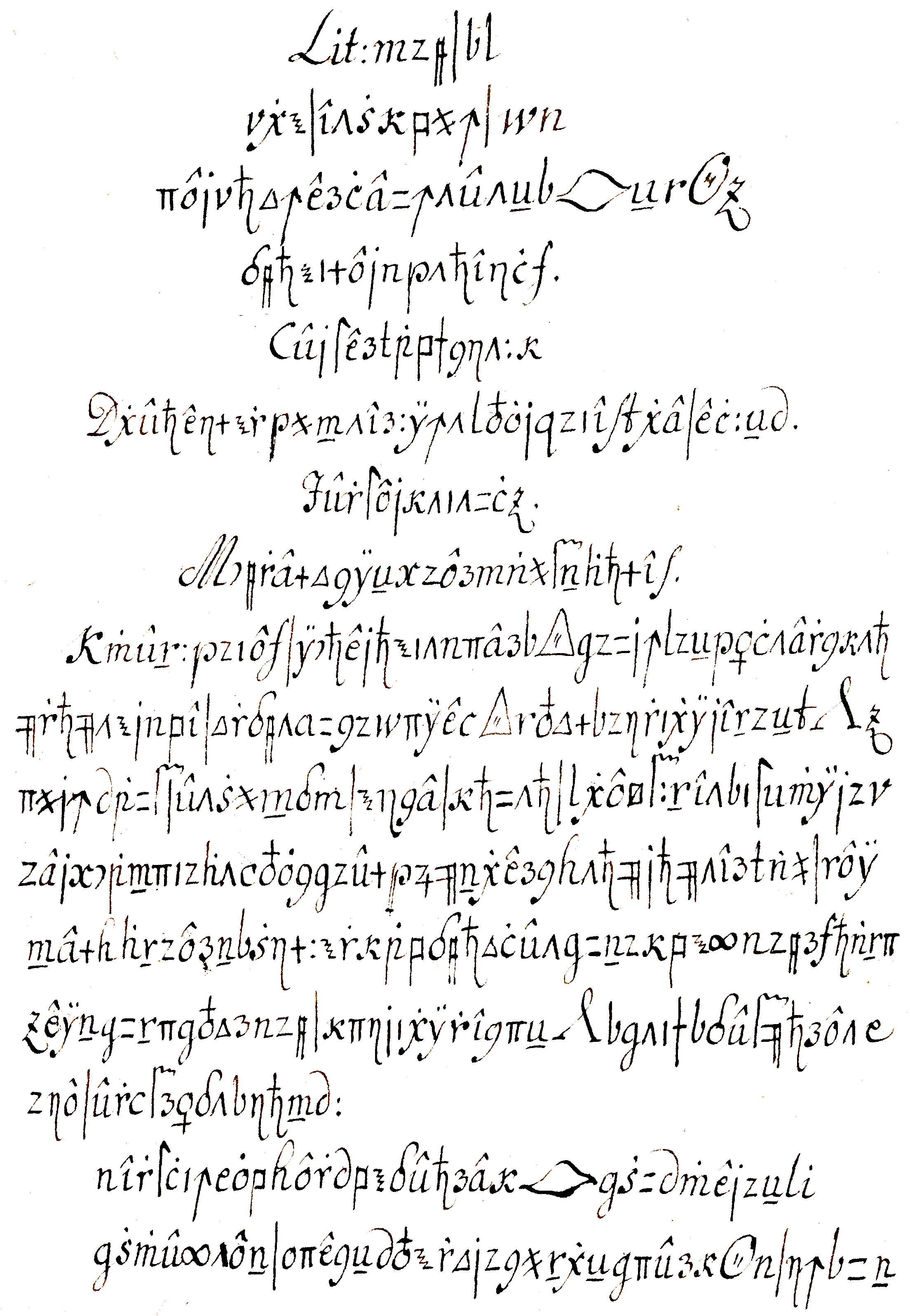}
            \caption[]{One page from Copiale cipher}
            \label{fig:one-page-copiale}
        \end{subfigure}
        \begin{subfigure}[b]{0.5\textwidth}   
            \centering 
            \includegraphics[width=0.8\textwidth]{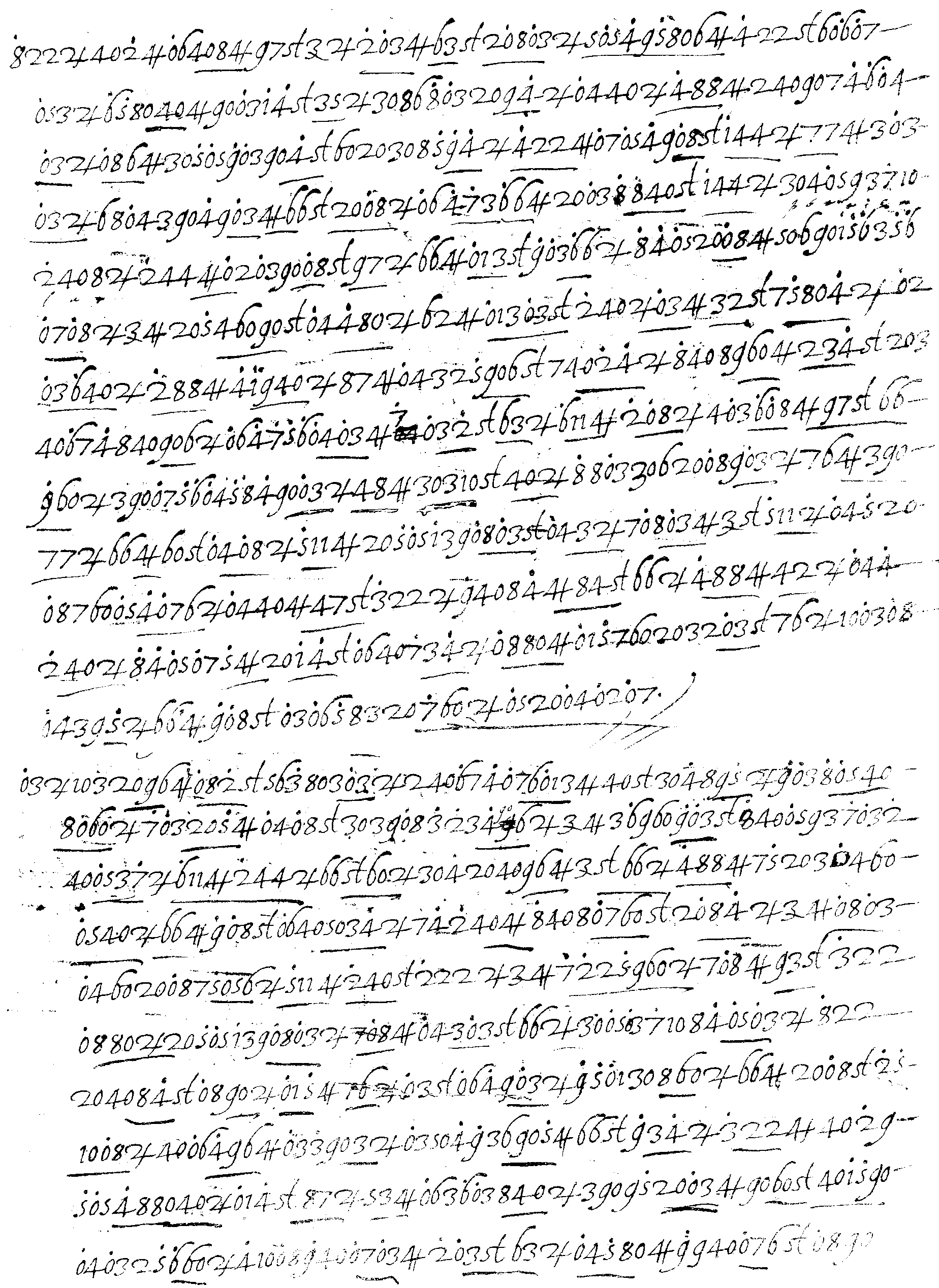}
            \caption[]{Spagna cipher}
            \label{fig:spagna}
        \end{subfigure}%
        \begin{subfigure}[b]{0.5\textwidth}   
            \centering 
            \includegraphics[width=0.8\textwidth]{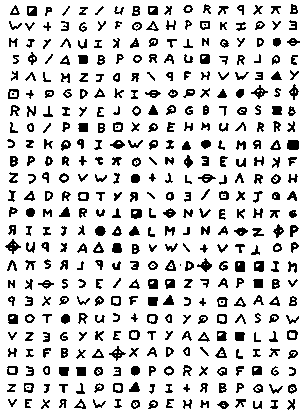}
            \caption[]{Zodiac cipher}
            \label{fig:zodiac}
        \end{subfigure}
        \caption{Pages from four cipher manuscripts.} 
        \label{fig:sample-manuscripts}
    \end{figure*}
    
\begin{figure*}  
\begin{center}
\begin{normalsize}
\begin{tabular}{|l|l|} \hline
Input cipher & \rule{0pt}{0.9cm}\includegraphics[height=0.7cm]{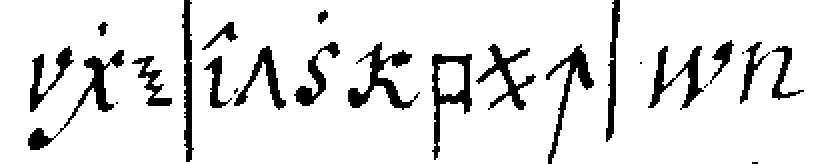}  \\ \hline
Segmentation  & \rule{0pt}{0.9cm}\includegraphics[height=0.7cm]{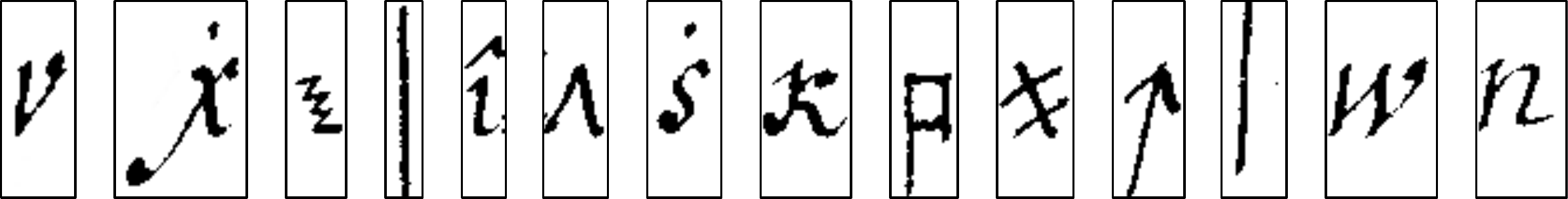} \\ \hline
Transcription &\verb| v x. zzz bar ih lam s.  k sqp ki arr bar  w  n|  \\ \hline
Decipherment &  \verb| _  g   e   s  e   t  z  _   b  u  ch   s  _  _|  \\ \hline
\end{tabular}
\end{normalsize}
\end{center}
\caption{Steps in decipherment a cipher manuscript.}
\label{fig:manual-transcript}
\end{figure*}

\section{Introduction}

European libraries and book collections are filled with undeciphered manuscripts dating from ca 1400 to 1900. These are often of historical significance, but historians cannot read them.  Over recent years, a large number of ciphers are being scanned, collected, and put online for experimentation \cite{histcorp}.  
Figure~\ref{fig:sample-manuscripts} shows examples of cipher manuscripts. These ciphers are all considerable low-resource datasets from aspects of using one-off alphabets and glyphs. From the fraction of manuscripts that have been deciphered, cipher systems include simple substitution, homophonic substitution (where there may be many ways to encipher a given plaintext letter), substitution-transposition, nomenclators (where symbols may stand for whole words), or a combination of those.  Plaintext languages include Latin, English, French, German, Italian, Portuguese, Spanish, Swedish, and so on.

Manual decipherment requires three major steps (Fig.~\ref{fig:manual-transcript}):

\begin{itemize}  
\item {\bf Segmentation}.  First, we decide where each character begins and ends.  Even though ciphers often employ novel one-off alphabets, human analysts are quite good at segmenting lines into individual characters.  However, problems do arise.  For example, in the Borg cipher (Figure~\ref{fig:sample-manuscripts}a), should \inlinegraphics{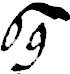} be segmented as one character or two? 
\item {\bf Transcription}.  Next, we convert the written characters into editable text, suitable for automatic analysis, such as character frequency counting.  As Figure~\ref{fig:manual-transcript} shows, this may involve inventing nicknames for characters that cannot be typed (e.g., {\em zzz} for \inlinegraphics{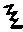}).  A human analyst can do this quite accurately, though mistakes happen.  For example, in the Copiale cipher, should \inlinegraphics{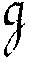} and \inlinegraphics{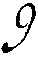} be transcribed the same way, or are they actually distinct ciphertext symbols?
\item {\bf Decipherment}.  Finally, we guess a cipher key that (when applied to the transcription) yields sensible plaintext.  This is the hardest step for a human analyst, requiring intuition, insight, and grunt work.
\end{itemize}

Segmentation and transcription can be performed either manually or by semi-automatically, with post-editing. Given the number of historical encrypted manuscripts, manual transcription is infeasible, because it is too time-consuming, expensive, and prone to errors \cite{Fornes_2017}. 

We would like to automate {\em all} of these steps, delivering a camera-phone decipherment app that a historian could use directly in the field. Automation efforts to date, however, have focused primarily on the decipherment step.  Therefore, the problem we attack in this paper is {\em automatic decipherment directly from scanned images}.\footnote{Our code is available at \newline \url{https://github.com/yinxusen/decipherment-images}}

Existing optical-character recognition (OCR) techniques are challenged by cipher manuscripts.  The vast bulk of modern handwritten OCR requires large supervised datasets \cite{Smith:2007:OTO:1304596.1304846,5539867,6065441}, whereas ciphers often use one-off alphabets for which no transcribed data exists. Back before supervised datasets were available, early OCR research proposed unsupervised identification and clustering of characters \cite{Nagy:1987:DSC:28809.28825,902858,ThomasOCR,4378705}. This is the general approach we follow here. Also in the unsupervised realm, recent work on historical documents focuses on printed, typeset texts \cite{conf/acl/Berg-KirkpatrickDK13,taylor14}.  Though these methods model various types of noise, including ink bleeds and wandering baselines, they expect general consistency in font and non-overlapping characters.  

The novel contributions of our paper are:

\begin{itemize}
\item Automatic algorithms for character segmentation, character clustering, and decipherment for handwritten cipher manuscripts.
\item Evaluation on image data from multiple ciphers, measuring accuracy of individual steps as well as end-to-end decipherment accuracy.
\item Improved techniques for joint inference, merging transcription and decipherment.
\end{itemize}

\section{Data}\label{sec:data}
We perform experiments on two historical manuscripts---Borg (Figure~\ref{fig:one-page-borg}) and Copiale (Figure~\ref{fig:one-page-copiale})---and two synthetic ciphers (see Table~\ref{tab:datasets}).
All images are black-and-white in PNG format. 

\begin{table}  
\centering
\begin{normalsize}
\begin{tabular}{|l|r|r|r|} \hline
Cipher & \# pages & \# characters & alphabet\\ 
 &  &  & size \\ 
\hline
Courier & 1 & 653 & 22 \\ \hline
Arial & 1 & 653 & 22 \\ \hline
Borg & 3 & 1054 & 23 \\ \hline
Copiale & 10 & 6491 & 79\\ \hline
\end{tabular}
\end{normalsize}
\caption{Statistics of cipher image datasets used in this paper.}\label{tab:datasets}
\end{table}

{\bf Borg.}
This is a 408-page manuscript from the 17th century, automatically deciphered by \cite{nada}. Its plaintext language is Latin.  A few pages of the Borg cipher contain plaintext Latin fragments, which we remove from the images. In our experiments, we choose three consecutive pages and trim margins of each page (872 x 1416 pixels on average).

{\bf Copiale.}
This is a 105-page cipher from the 18th century, deciphered by \cite{Knight:2011:CC:2024236.2024239}. The plaintext of Copiale is German. Copiale uses eight nomenclators that map to whole plaintext words, so we remove them from cipher images. In our experiments, we use the first 10 pages (1160x1578 pixels on average). 

{\bf Courier} (synthetic). We encipher a 653-character English text (simple substitution, no space), print it with fixed-width Courier font, then scan it into a PNG file (1956x2388 pixels).

{\bf Arial} (synthetic).  We create a similar image using variable-width Arial font (1976x2680 pixels).

We now turn to automatic methods for segmenting, transcribing, and deciphering.

\section{Automatic Segmentation}\label{sec:segmentation}


We define the upper-left corner of a page image as its origin, and the upper boundary as x-axis, left boundary as y-axis.

\begin{enumerate}[label=(\subscript{T}{\arabic*})]
\item We draw horizontal lines $y=c$, to split the manuscript into rows of characters;
\item We then draw vertical lines $x=c$,  slant lines $y=bx$, or cubic curves $y=ax^3+bx+c$ 
in each row of image to split characters.
\end{enumerate}

Taking Task $T_2$ as an example, for an image row with $m$ characters, we need to find cutting points $c_1, c_2, \cdots, c_m$ on the x-axis to draw curves.\footnote{The last cutting point $c_m$ is unnecessary, we use it to write clear equation.} Cutting points and curves we choose on the x-axis should meet the following (conflicting) requirements:

\begin{figure}
\begin{center}
\includegraphics[width=0.4\textwidth]{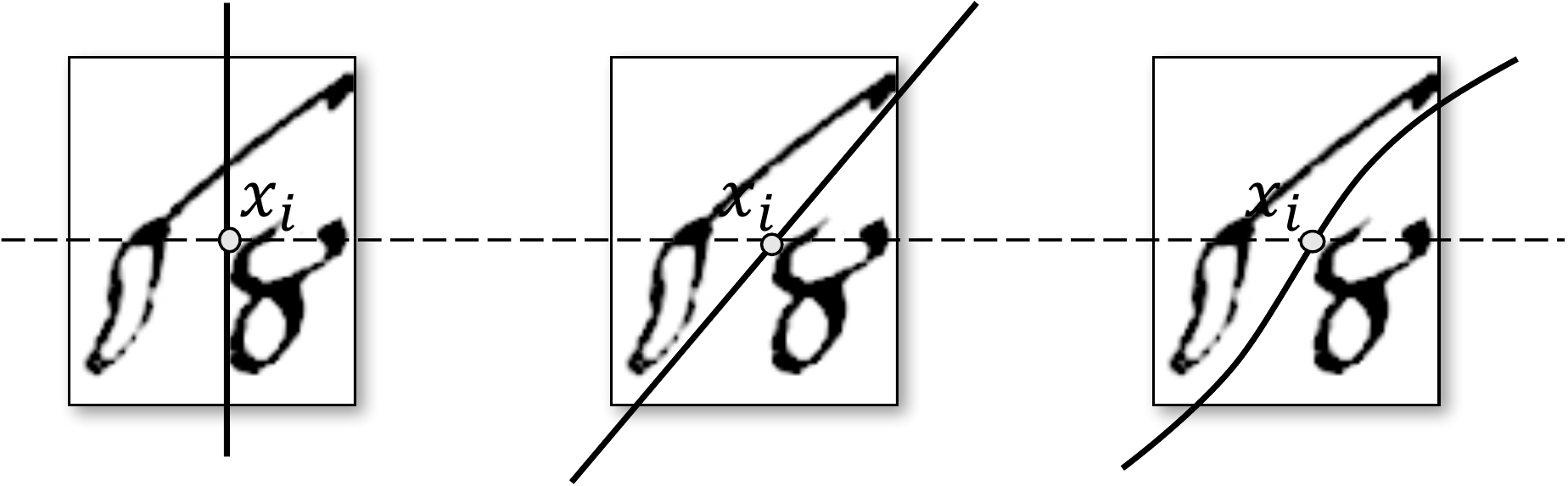}
\end{center}
\caption{Vertical, slant, and cubic character-segmentation curves in solid lines cutting through the same point $x_i$.
In this example, we choose the slant line as our cutting curve, and the number of intersected black pixels is $b_i=0$.}
\label{fig:mixed-curves}
\end{figure}

\begin{enumerate}[label=(\subscript{R}{\arabic*})]
\item the number of cutting points should be $m$.
\item the widths of characters should be as similar as possible.
\item curves drawing across cutting points should intersect with as few black pixels as possible.
\end{enumerate}


We use a generative model to formulate the requirements. At every point $x_i$ on the x-axis of a row image, we use a set of pre-defined curves to cut through it (see Figure~\ref{fig:mixed-curves}), and choose one with the minimum intersected black pixels $b_i$. When curves tie, we choose the simplest curve. If the row has $W$ total pixel columns, our observed data is the sequence of black pixel numbers $b_1, b_2, \cdots, b_W$ collected from all cutting points on the x-axis $x_1, x_2, \cdots, x_W$. Our goal is to choose $m$ cutting points out of $x_1, x_2, \cdots, x_W$ to meet the requirements.

The generative story is first to choose the number of characters $m$ of the row image according to a Gaussian distribution $m\sim P_{\phi_1, \sigma_1}(m)$. Then starting from the beginning of the row image, we generate the width of the next character from another Gaussian distribution $w_i \sim P_{\phi_2, \sigma_2}(w_{i})$.\footnote{Note that according to $R_2$, given $\phi_1$, we have $\phi_2 = W / \phi_1$. So we can omit $\phi_2$ in practice.} Subsequently we use a geometric distribution $p$ to generate the ``observed'' number of black pixels $b_i \sim P_{p}(b_{i})$. We repeat for all $m$ characters. We manually set parameters of the three distributions $\phi_1, \sigma_1, \sigma_2, p$ for each cipher. 

We use Viterbi decoding to find the sequence $c_1, c_2, \cdots, c_m$ that best satisfies $R_1$, $R_2$, and $R_3$.

\[\argmax_{m, c_1 \cdots c_m} P_{\phi_1, \sigma_1}(m)\prod_{i=1}^{m}P_{\sigma_2}(w_i)P_{p}(b_{i})\]


\begin{figure*}    
\centering
        \begin{subfigure}[b]{0.5\textwidth}
        \centering
                \includegraphics[width=0.9\textwidth]{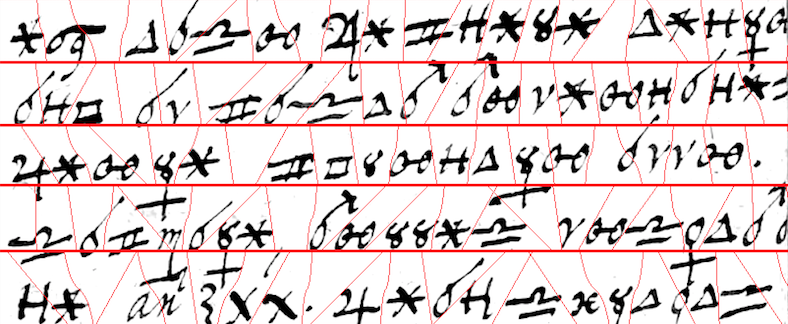}
                \label{fig:gull}
        \end{subfigure}%
        \begin{subfigure}[b]{0.5\textwidth}
        \centering
                \includegraphics[width=0.9\textwidth]{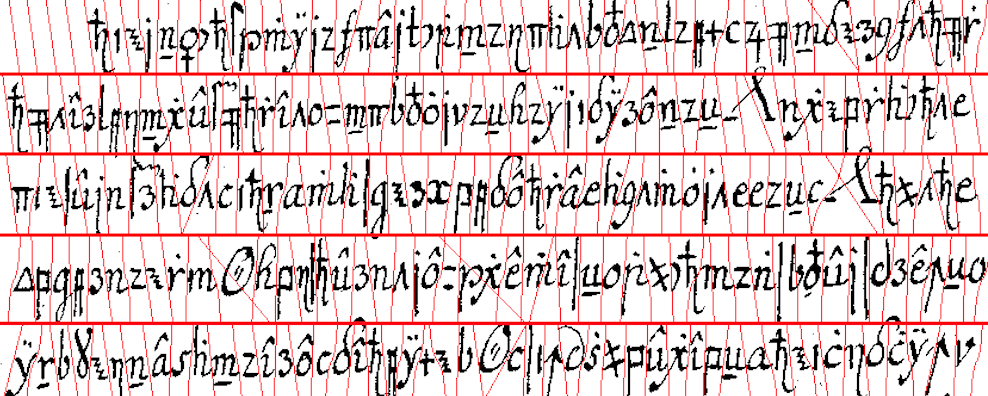}
                \label{fig:gull2}
        \end{subfigure}
        \caption{Results of automatic segmentation for part of Borg (left) and Copiale (right). We first use horizontal lines to segment rows, then use curves to segment characters in each row.}\label{fig:segmentation-examples}
\end{figure*}

Figure~\ref{fig:segmentation-examples} shows automatic segmentation results on snippets of manuscripts shown in Figure~\ref{fig:one-page-borg} and Figure~\ref{fig:one-page-copiale}. 

We manually create gold segmentations by cropping characters with a mouse.  However, we evaluate our segmenter only as it contributes to accurate transcription. 


\section{Transcription}\label{sec:transcription}



For transcription, we:

\begin{enumerate}
\item Scale all character images to 105x105 pixels.
\item Convert each character image into a low-dimensional feature-vector representation.
\item Cluster feature vectors into similar groups.  
\item Output a sequence of cluster IDs.
\end{enumerate}



We implement two methods to transform a character image $x$ into a fixed-length feature vector $g$ with $g = F(x)$.

First, we propose a pairwise similarity matrix (SimMat). Given a sequence of character images $X = \{x_1, x_2, \cdots, x_n\}$, SimMat computes the similarity between every pair of images as  $s(x_i, x_j)$. Image $x_i$ is then transformed into a n-dim vector according to the following equation,

\[F_{SimMat}(x_i; X)= [s(x_i, x_j)]_{j=1}^n \]

The similarity function $s(x_i, x_j)$ is the maximum of cross correlate matrix of two images. We use the \texttt{signal.correlate2d} function in Scipy package.

\begin{figure}
\begin{center}
\includegraphics[width=0.475\textwidth]{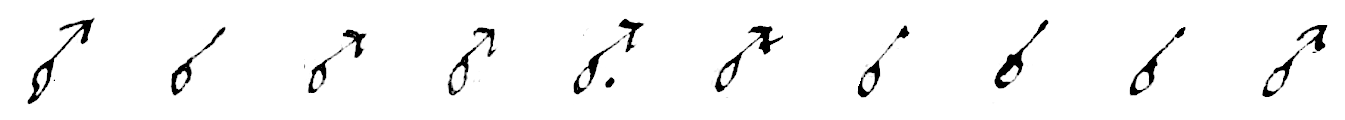}
\end{center}
\caption{Part of a cluster from the Borg dataset using SimMat features. Here, two different cipher symbols are conflated into a single cluster.}

\label{fig:cluster-3}
\end{figure}

SimMat is a non-parametric feature extractor with a $O(n^2)$ time complexity, which makes it hard to apply to long ciphers.
A sample cluster is shown in Figure~\ref{fig:cluster-3}.

Our second strategy exploits Omniglot, a labeled character-image dataset \cite{DBLP:conf/cogsci/LakeSGT11},
containing 50 different alphabets, about 25 unique characters each, and 20 handwritten instances per. 

\begin{enumerate}
\item We follow \cite{Koch2015SiameseNN} to train a Siamese Neural Network (SNN) on pairs of Omniglot images.  The SNN  outputs 0 if  two input images represent the same character.
\item We feed cipher character images into the SNN to extract feature representations.
\end{enumerate}

\begin{figure}
\begin{center}
\includegraphics[width=0.475\textwidth]{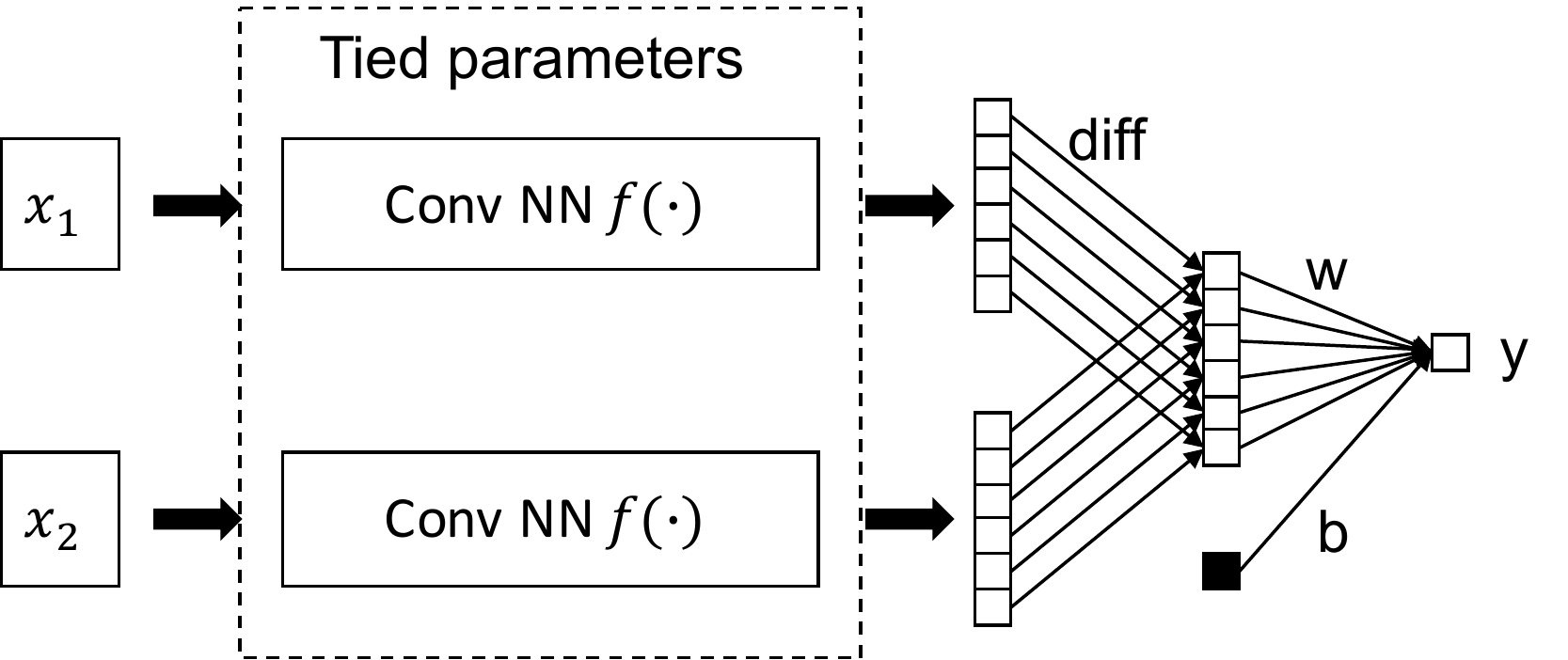}
\end{center}
\caption{The architecture of Siamese Neural Network. $x_1$ and $x_2$  are a pair of character image inputs. 
Output $y=0$ means $x_1$ and $x_2$ represent the same character type.}
\label{fig:diag-snn}
\end{figure}

The SNN architecture  \cite{Koch2015SiameseNN} is shown in Figure~\ref{fig:diag-snn}. The SNN has a visual feature extraction part $f(\cdot)$, which is a convolution neural network, plus a single-layer neural classifier. Given an input image pair $(x_1, x_2)$, the output is

\noindent $y = sigmoid(w*(f(x_1) - f(x_2)) + b)$.

We turn the classifier into a feature extractor by removing its classification part:

\noindent $F_{SNN}(x_i) = w*f(x_i)$.



For clustering feature vectors, we use a standard Gaussian mixture model (GMM).\footnote{We set GMM covariance type as diagonal, spherical, and fix-cov=$\{1, 0.1, 0.01, 0.001\}$ and choose the best one for each dataset.}  Finally, as our transcription, we output a sequence of cluster IDs.

\begin{table*}    
\centering
\begin{normalsize}
\begin{tabular}{| l| l|}
\hline
Gold transcription & \verb| z  o  d  i  a  c  k  i  l  l  e  r |\\ 
Automatic transcription & \verb|c0 c1 c2 c3 c3 c4 c5 c3 c6 c6 c7 c8 | \\
\hline
Best alphabet mapping & c0 - z, c1 - o, c2 - d, c3 - i, c4 - c, c5 - k, c6 - l, c7 - e, c8 - r \\
 Edit distance after mapping & 1 (a - c3) \\
 NEDoA & 1 / len(zodiackiller) = 1 / 12 = 0.083 \\ \hline
\end{tabular}
\end{normalsize}
\caption{Computing automatic transcription accuracy using the NEDoA metric. We map  cluster ID types to gold transcription symbol types, make substitutions on the automatic transcription, then compute edit distance with gold. We search for the mapping that leads to the minimum normalized edit distance.}\label{tbl:NEDoA}
\end{table*}



{\bf Evaluating Automatic Transcription.}
We manually create gold-standard transcriptions for all our ciphers.  To judge the accuracy of our automatic transcription, we cannot simply use edit distance, because cluster IDs types do match human-chosen transcription symbols. Therefore, we map cluster ID types into human transcription symbols (many to one), transform the cluster IDs sequence accordingly, and then compute normalized edit distance.  There are many possible mappings---we choose the one that results in the minimal edit distance.  We have two methods to accomplish this, one based on integer-linear programming, and one based on expectation-maximization (EM).  We call this {\bf Normalized Edit Distance over Alignment} (NEDoA).  Table~\ref{tbl:NEDoA} gives an example.

Table~\ref{tbl:image-seg} compares NEDoA transcription accuracies under the SimMat and SNN feature extractors.  SNN outperforms SimMat.  The table also compares transcriptions from gold segmentation and automatic segmentation.  Automatic segmentation on Borg degrades transcription accuracy.



\begin{table}  
\centering
\begin{normalsize}
\begin{tabular}{|l|r|r|r|}
\hline
   & \multicolumn{2}{r|}{\begin{tabular}[c]{@{}l@{}}Gold segmentation\end{tabular}} & \begin{tabular}[r]{@{}l@{}}Auto-\\segmentation\end{tabular} \\ \hline
\begin{tabular}[r]{@{}l@{}}Feature \\ extractor\end{tabular} & SimMat   & SNN & SNN  \\ \hline
Courier  & 0.03 & 0.03 & 0.03 \\ \hline
Arial  & 0.03 & 0.04 & 0.08 \\ \hline
Borg  & 0.63 & 0.22  & 0.57  \\ \hline
Copiale & 0.87 & 0.37  & 0.44  \\ \hline
\end{tabular}
\end{normalsize}
\caption{Automatic transcription error rates (using NEDoA).}  
\label{tbl:image-seg}
\end{table}

\section{Decipherment from Transcription}\label{sec:decipher-from-transcription}

We can decipher from auto-transcription with the noisy channel model \cite{Knight99acomputational}.


\begin{figure}
\begin{center}
\includegraphics[width=0.45\textwidth]{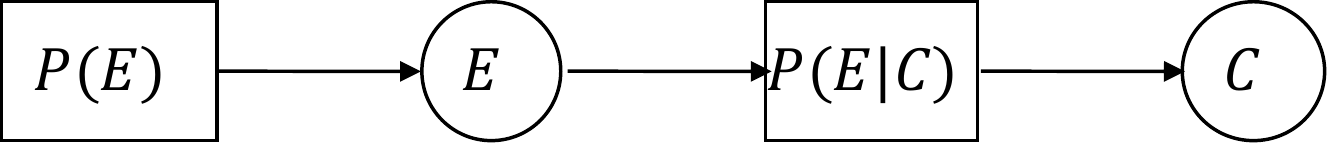}
\end{center}
\caption{Noisy channel model for decipherment. $P(E)$ is a character language model and $P(E|C)$ is a channel model.}
\label{fig:diag-deciphering}
\end{figure}

This generative model (Figure~\ref{fig:diag-deciphering}) first creates a sequence of plaintext characters $E = e_1e_2\cdots e_n$ with a character n-gram model, then uses a channel model $P(C|E)$, transforms $E$ into cipher text $C=c_1c_2\cdots c_n$ character-by-character. The probability of our observation $C$ is

\[P(C) = \sum_E P(E) P(C|E)\]

We can find the optimal channel model with the EM algorithm:

\[
P(C|E) = \argmax_{P(C|E)} P(C) 
\]

After we get the trained channel model, we use Viterbi decoding to find out the plaintext:

\[
E = \argmax_{E} P(E|C) \propto \argmax_{E} P(E)P(C|E) 
\]

We call the first step \textbf{deciphering}, and the second step \textbf{decoding}. Combining segmentation, transcription, and decipherment, we create a pipeline to decipher from a scanned image, which we call \textbf{3-stage decipherment}.

{\bf Results for 3-stage Decipherment.} We build pre-trained bigram character language models of English, Latin, and German. Since our cipher datasets have been deciphered, for each dataset we generate gold plaintext from gold transcription. We evaluate decipherment with normalized edit distance (NED).

Table~\ref{tbl:3-stage-decipher} compares decipherment error rates under gold segmentation and automatic segmentation.  We also study decipherment under gold transcription.  Our fully automatic system deciphers Copiale at 0.51 character error.  While high, this is actually remarkable given that our transcription has 0.44 error.  It seems that our fully-connected noisy channel decipherment model is able to overcome transcription mistakes by mapping the same cluster ID onto different plaintext symbols, depending on context.


Even so, we notice substantial degradation along the pipeline.  Human analysts also revise transcriptions once decipherments are found.

\begin{table}[!htb]
\centering
\begin{normalsize}
\begin{tabular}{|l|r|r|r|}
\hline
\multirow{2}{*}{} & \multicolumn{2}{l|}{Auto-transcription} & \multirow{2}{*}{\begin{tabular}[c]{@{}l@{}}Gold \\ transcription\end{tabular}} \\ \cline{2-3}
                     & Gold seg          & Auto-seg          &                                                                                         \\ \hline
Courier              & 0.08              & 0.08              & 0.08                                                                                    \\ \hline
Arial                & 0.07              & 0.16              & 0.08                                                                                    \\ \hline
Borg                 & 0.35              & 0.76              & 0.01                                                                                    \\ \hline
Copiale              & 0.46              & 0.51              & 0.20                                                                                    \\ \hline
\end{tabular}
\end{normalsize}{}
\caption{Error rates of 3-stage decipherment (NED) based on gold segmentation and auto-segmentation, compared with deciphering from gold transcription.}
\label{tbl:3-stage-decipher}
\end{table}

\section{Decipherment from Character Images}\label{sec:lm-gmm}

We propose {\bf 2-stage decipherment}, which models transcription and decipherment as a single integrated step. 


\subsection{Language Model Constrained Gaussian Mixture Model (LM-GMM)}

\begin{figure}
\begin{center}
\includegraphics[width=0.45\textwidth]{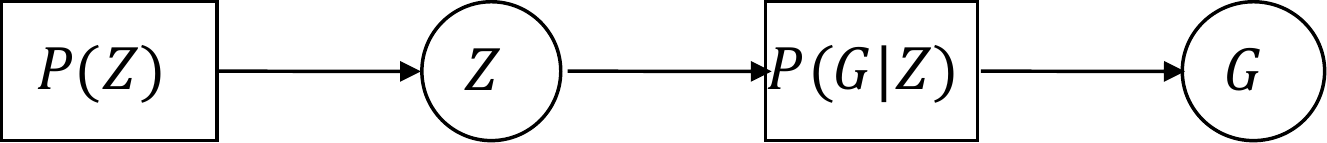}
\end{center}
\caption{GMM. $P(Z)$ is a multinomial distribution over cluster IDs, and $P(G|Z)$ is a mixture of Gaussian distributions.}
\label{fig:diag-gmm}
\end{figure}

{\bf GMM.} Given a sequence of feature vectors $G=\{g_1, g_2, \cdots, g_n\}$ generated by the SNN feature extractor, GMM generates $G$ by first using a multinomial distribution $P(Z)$ to choose cluster assignments, then uses the mixture of Gaussian distributions $P(G|Z)$ to generate feature vectors.
, as shown in Figure~\ref{fig:diag-gmm}.

\[P(G) = \sum_Z P(Z)P(G|Z)\]

\begin{figure*}
\begin{center}
\includegraphics[width=0.7\textwidth]{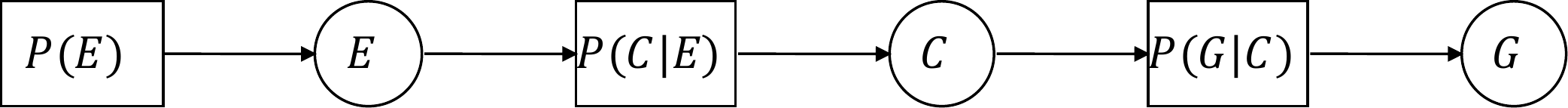}
\end{center}
\caption{LM-GMM. $P(E)$ is character language model, $P(C|E)$ is a channel model, and $P(G|C)$ is a mixture of Gaussian distributions.}
\label{fig:diag-lm-gmm}
\end{figure*}

\begin{figure}
\begin{center}
\includegraphics[width=0.45\textwidth]{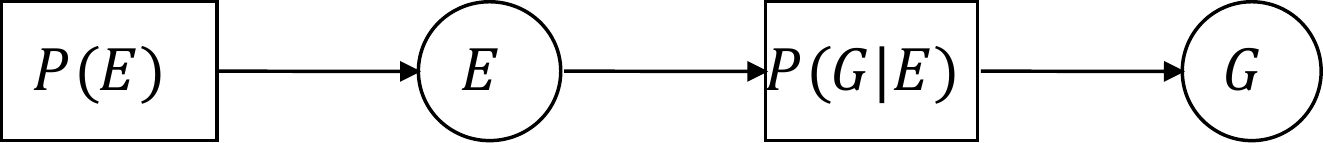}
\end{center}
\caption{Simplified LM-GMM for simple substitution ciphers. }
\label{fig:diag-simple-lm-gmm}
\end{figure}

\begin{figure*}  
        \begin{subfigure}[b]{0.475\textwidth}
                \includegraphics[width=\linewidth]{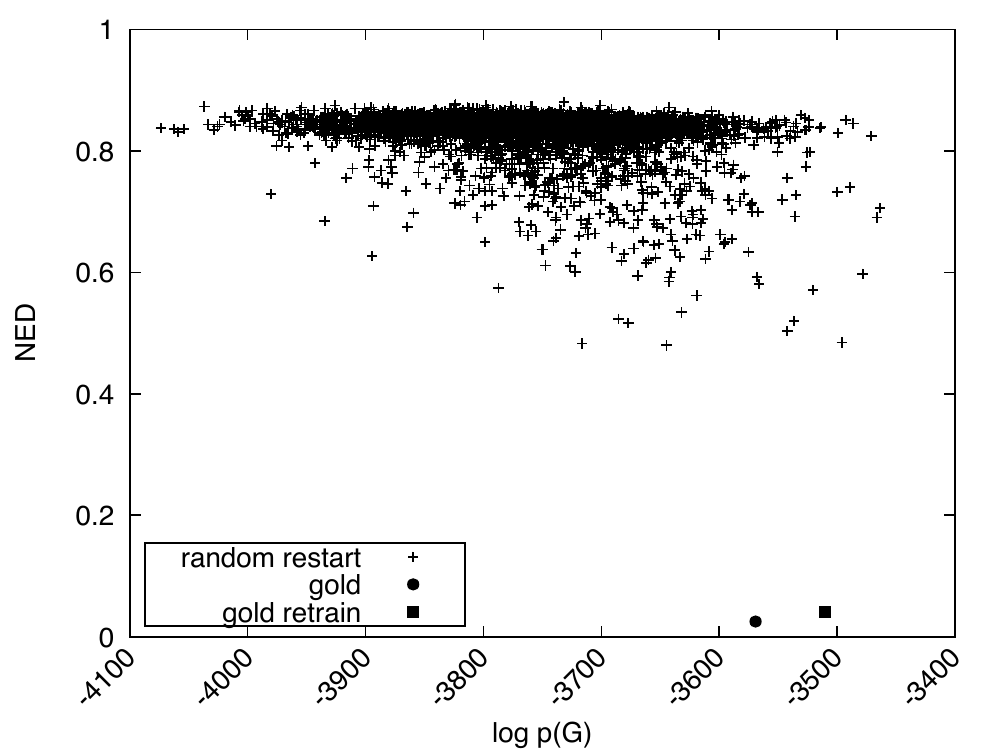}
                \label{fig:small-eb1}
        \end{subfigure}%
        \hfill
        \begin{subfigure}[b]{0.475\textwidth}
                \includegraphics[width=\linewidth]{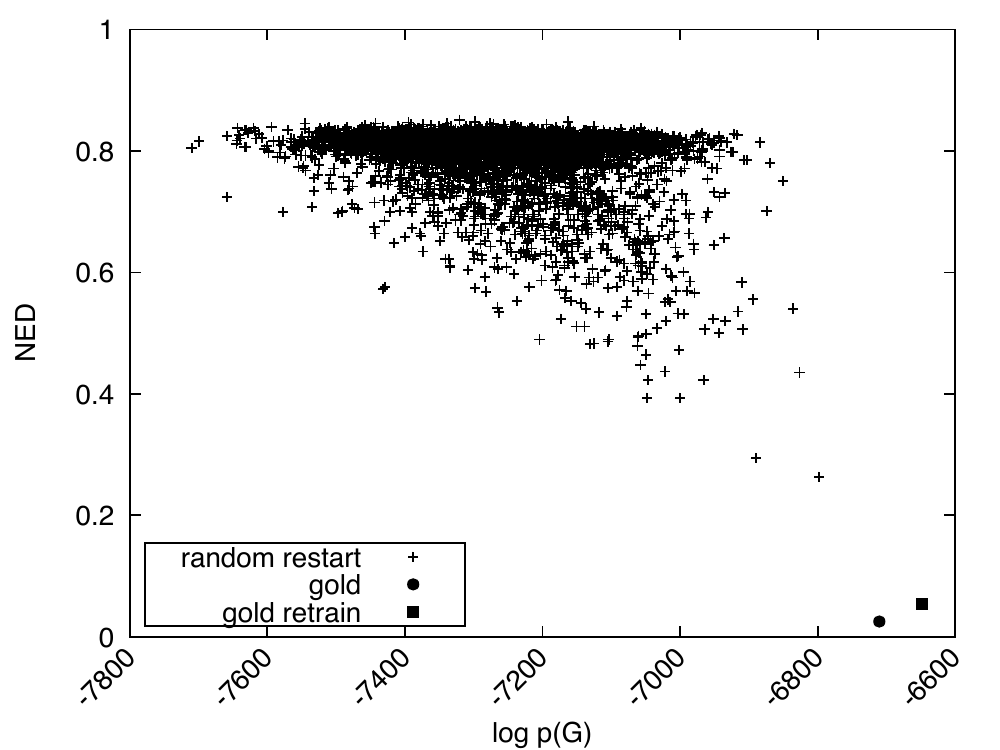}
                \label{fig:small-eb3}
        \end{subfigure}%
        \caption{5000 random restarts of training LM-GMM (left), and LM-GMM with cubic LM (right) on Courier dataset marked with plus (+). X-axis is the likelihood of the Courier dataset according to LM-GMM, y-axis is NED between deciphered text and gold plaintext. The solid dot is the gold model, and the solid square is the LM-GMM training result initialized from the gold model.}\label{fig:em-restarts}
\end{figure*}

{\bf LM-GMM.} Instead of using the multinomial distribution $P(Z)$ to choose clusters, LM-GMM uses the decipherment language model to choose appropriate character sequences. Since we do not have cipher language model, we use the noisy channel model as the {\bf cipher language model} as shown in Figure~\ref{fig:diag-lm-gmm}.

\[P(G) = \sum_E P(E)\sum_C P(C|E)P(G|C)\]


{\bf Simplified LM-GMM.} LM-GMM can be simplified for simple substitution ciphers. Since simple substitution ciphers use one-to-one and onto mappings between plaintext alphabet and cipher alphabet, the channel model $P(C|E)$ is not necessary. We imagine, for example, that Borg is written in Latin, but the author writes Latin characters strangely. The simplified model is

\[P(G) = \sum_E P(E)P(G|C)\]

\subsection{LM-GMM Model Error}\label{sec:model-problem}

LM-GMM is a combination of discrete distributions (LM, channel) and a continuous one (GMM). 
Figure~\ref{fig:em-restarts} (left) shows results of 5,000 random restarts (+).
The x-axis is the log-likelihood of observed feature vectors. The y-axis is decipherment NED. We also plot the gold model by generating the Gaussian Mixture $P(G|C)$ with gold plaintext, as the solid dot. Training from the gold model, we can reach the solid square. 


The gold model does not receive the highest model score.  This modeling error is caused by the strong GMM multiplying with the character language model---we tend to choose the result satisfying the GMM part.  To fix this, we use $P(G)=\sum_E P(E)^3P(G|C)$ to highlight the importance of the language model during deciphering phase, leading to a result shown in Figure \ref{fig:em-restarts} (right).

\subsection{LM-GMM Search Error}\label{sec:init-problem}

Now we observe that even after many EM restarts, we cannot reach a model that scores as well as the gold model.



\begin{figure}
\begin{center}
\includegraphics[width=0.475\textwidth]{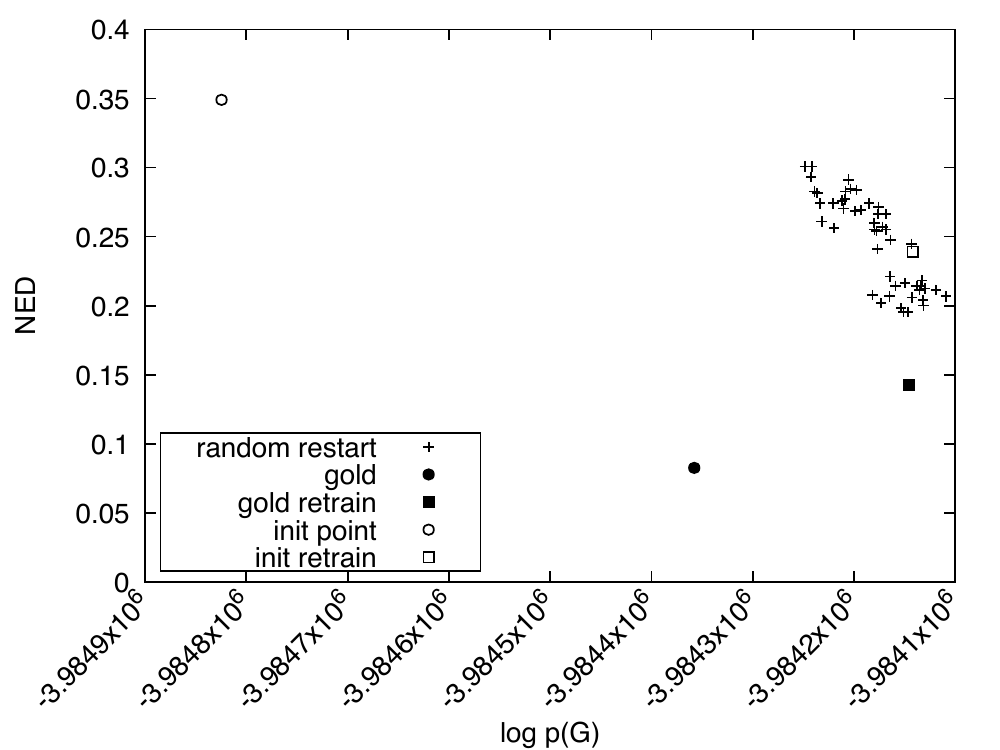}
\end{center}
\caption{50 restarts of LM-GMM on Borg dataset initialized from 3-stage decipherment with random noise (+). The x-axis is the log-likelihood of Borg vectors according to LM-GMM, and the y-axis is NED between deciphered text and gold plaintext. The hollow circle is the 3-stage decipherment result mapped onto this plot, and the hollow square is the LM-GMM training result initialized from the hollow circle. The solid dot is the gold model, and the solid square is the LM-GMM training result starting from the solid dot.}
\label{fig:borg-init-better}
\end{figure}

To fix this search problem, we randomly restart our EM training from a GMM model computed from plaintext from 3-stage training. 
We illustrate the initialization method on the Borg dataset (Figure~\ref{fig:borg-init-better}).  The initialization point comes from 3-stage decipherment (NED=0.35), which 2-stage decipherment improves to NED=0.20.  This approaches the retrained gold model (NED=0.15), with only 50 restarts.

\subsection{Evaluation of 2-stage Decipherment}

Decipherment results are shown in Table~\ref{tbl:deciphering-results}, which compares 2- and 3-stage decipherment under auto- and gold segmentation. All results are trained with bigram character language model. Results of 2-stage decipherment use the initialization method described above. From the results we can see that 2-stage decipherment outperforms 3-stage decipherment, especially for Borg and Copiale.

\begin{table}
\centering
\begin{normalsize}

\begin{tabular}{|l|r|r|r|l|}
\hline
\multirow{2}{*}{} & \multicolumn{2}{r|}{Gold seg} & \multicolumn{2}{r|}{Auto-seg} \\ \cline{2-5} 
                     & 3-stage            & 2-stage           & 3-stage            & 2-stage           \\ \hline
Courier              & 0.08               & 0.06              & 0.08               & 0.06              \\ \hline
Arial                & 0.07               & 0.04              & 0.16               & 0.11              \\ \hline
Borg                 & 0.35               & 0.20              & 0.76               & 0.69              \\ \hline
Copiale              & 0.46               & 0.41              & 0.51               & 0.50              \\ \hline
\end{tabular}
\end{normalsize}{}
\caption{Error rates of 2-stage decipherment (NED) on both gold segmentation and auto-segmentation, compared with 3-stage decipherment.}
\label{tbl:deciphering-results}
\end{table}

Instead of using bigram language model, we also use a trigram language model on Borg, as shown in Table~\ref{tbl:higher-order}. Both 3-stage and 2-stage decipherment get better results.

\begin{table}
\centering
\begin{normalsize}

\begin{tabular}{|l|r|r|}
\hline
& 3-stage & 2-stage \\
 & decipherment & decipherment \\ \hline
Bigram     & 0.35   & 0.20   \\ \hline
Trigram   & 0.24   & 0.16  \\ \hline
\end{tabular}
\end{normalsize}{}
\caption{Decipherment error rates (NED) with bigram and trigram language models for Borg, using gold segmentation.}
\label{tbl:higher-order}
\end{table}

\section{Conclusion and Future Work}\label{conclude}

In this paper, we build an end-to-end system to decipher from manuscript images.
We show that the SNN feature extractor with a Gaussian mixture model can be good for unseen character clustering. We fix our EM search problem for LM-GMM by using a better initialization method.

Interesting future work can include 1-stage decipherment. Can we use our cipher language model to improve the image segmentation? Can we merge image segmentation into the whole EM training framework? How much benefit can we get? 

Finally, to realize a fully-automatic camera-phone decipherment app, we need to lift several assumptions we made in the paper.  These include knowing the plaintext language and cipher system \cite{Nuhn2014CipherType,Hauer2016DecodingAT,nada}, pre-processing images to remove margins and non-cipher text, knowing the cipher alphabet size, and cipher-specific setting of segmentation parameters.

\section*{Acknowledgments}
The authors thank all people who have deciphered these manuscripts to make this work possible. This work was sponsored by DARPA (grant HR0011-15-C-0115), by the Swedish Research Council (grant E0067801), by the USC Annenberg Graduate Fellowship, and by a gift from Google, Inc.



\bibliographystyle{IEEEtran}
\bibliography{IEEEabrv,bare_conf}

\begin{thebibliography}{10}
\providecommand{\url}[1]{#1}
\csname url@samestyle\endcsname
\providecommand{\newblock}{\relax}
\providecommand{\bibinfo}[2]{#2}
\providecommand{\BIBentrySTDinterwordspacing}{\spaceskip=0pt\relax}
\providecommand{\BIBentryALTinterwordstretchfactor}{4}
\providecommand{\BIBentryALTinterwordspacing}{\spaceskip=\fontdimen2\font plus
\BIBentryALTinterwordstretchfactor\fontdimen3\font minus
  \fontdimen4\font\relax}
\providecommand{\BIBforeignlanguage}[2]{{%
\expandafter\ifx\csname l@#1\endcsname\relax
\typeout{** WARNING: IEEEtran.bst: No hyphenation pattern has been}%
\typeout{** loaded for the language `#1'. Using the pattern for}%
\typeout{** the default language instead.}%
\else
\language=\csname l@#1\endcsname
\fi
#2}}
\providecommand{\BIBdecl}{\relax}
\BIBdecl

\bibitem{histcorp}
E.~Pettersson and B.~Megyesi, ``The histcorp collection of historical corpora
  and resources,'' in \emph{Proceedings of the Third Conference on Digital
  Humanities in the Nordic Countries}, 2018.

\bibitem{Fornes_2017}
A.~Forn\'{e}s, B.~Megyesi, and J.~Mas, ``Transcription of encoded manuscripts
  with image processing techniques,'' in \emph{Proceedings of Digital
  Humanities}, 2017.

\bibitem{Smith:2007:OTO:1304596.1304846}
R.~Smith, ``An overview of the tesseract {OCR} engine,'' in \emph{Proceedings
  of the Ninth International Conference on Document Analysis and Recognition},
  2007, pp. 629--633.

\bibitem{5539867}
A.~Kae, G.~Huang, C.~Doersch, and E.~Learned-Miller, ``Improving
  state-of-the-art {OCR} through high-precision document-specific modeling,''
  in \emph{IEEE Conference on Computer Vision and Pattern Recognition}, June
  2010, pp. 1935--1942.

\bibitem{6065441}
V.~Kluzner, A.~Tzadok, D.~Chevion, and E.~Walach, ``Hybrid approach to adaptive
  {OCR} for historical books,'' in \emph{IEEE International Conference on
  Document Analysis and Recognition}, Sept 2011, pp. 900--904.

\bibitem{Nagy:1987:DSC:28809.28825}
\BIBentryALTinterwordspacing
G.~Nagy, S.~C. Seth, and K.~Einspahr, ``Decoding substitution ciphers by means
  of word matching with application to {OCR},'' \emph{IEEE Transactions on
  Pattern Analysis and Machine Intelligence}, vol.~9, no.~5, pp. 710--715, May
  1987. [Online]. Available: \url{http://dx.doi.org/10.1109/TPAMI.1987.4767969}
\BIBentrySTDinterwordspacing

\bibitem{902858}
T.~K. Ho and G.~Nagy, ``{OCR} with no shape training,'' in \emph{Proceedings of
  15th International Conference on Pattern Recognition}, vol.~4, 2000, pp.
  27--30.

\bibitem{ThomasOCR}
T.~M. Breuel, ``Modeling the sample distribution for clustering {OCR},'' in
  \emph{Proc. SPIE, Document Recognition and Retrieval VIII}, vol. 4307, 12
  2000, pp. 193--200.

\bibitem{4378705}
G.~Huang, E.~Learned-Miller, and A.~McCallum, ``Cryptogram decoding for ocr
  using numerization strings,'' in \emph{Ninth International Conference on
  Document Analysis and Recognition}, vol.~1, Sept 2007, pp. 208--212.

\bibitem{conf/acl/Berg-KirkpatrickDK13}
T.~Berg-Kirkpatrick, G.~Durrett, and D.~Klein, ``Unsupervised transcription of
  historical documents.'' in \emph{Proceedings of the 51st Annual Meeting of
  the Association for Computational Linguistics}, 2013, pp. 207--217.

\bibitem{taylor14}
\BIBentryALTinterwordspacing
T.~Berg{-}Kirkpatrick and D.~Klein, ``Improved typesetting models for
  historical {OCR},'' in \emph{Proceedings of the 52nd Annual Meeting of the
  Association for Computational Linguistics}, 2014, pp. 118--123. [Online].
  Available: \url{http://aclweb.org/anthology/P/P14/P14-2020.pdf}
\BIBentrySTDinterwordspacing

\bibitem{nada}
N.~Aldarrab, ``Decipherment of historical manuscripts,'' Master's thesis,
  University of Southern California, Los Angeles, California, 2017.

\bibitem{Knight:2011:CC:2024236.2024239}
\BIBentryALTinterwordspacing
K.~Knight, B.~Megyesi, and C.~Schaefer, ``The copiale cipher,'' in
  \emph{Proceedings of the 4th Workshop on Building and Using Comparable
  Corpora: Comparable Corpora and the Web}, 2011, pp. 2--9. [Online].
  Available: \url{http://dl.acm.org/citation.cfm?id=2024236.2024239}
\BIBentrySTDinterwordspacing

\bibitem{DBLP:conf/cogsci/LakeSGT11}
\BIBentryALTinterwordspacing
B.~M. Lake, R.~Salakhutdinov, J.~Gross, and J.~B. Tenenbaum, ``One shot
  learning of simple visual concepts,'' in \emph{Proceedings of the 33th Annual
  Meeting of the Cognitive Science Society}, 2011. [Online]. Available:
  \url{https://mindmodeling.org/cogsci2011/papers/0601/index.html}
\BIBentrySTDinterwordspacing

\bibitem{Koch2015SiameseNN}
G.~Koch, R.~Zemel, and R.~Salakhutdinov, ``Siamese neural networks for one-shot
  image recognition,'' in \emph{Proceedings of the 32nd International
  Conference on Machine Learning}, 2015.

\bibitem{Knight99acomputational}
K.~Knight and K.~Yamada, ``A computational approach to deciphering unknown
  scripts,'' in \emph{Proceedings of the ACL Workshop on Unsupervised Learning
  in Natural Language Processing}, 1999.

\bibitem{Nuhn2014CipherType}
M.~Nuhn and K.~Knight, ``Cipher type detection,'' in \emph{Proceedings of the
  2014 Conference on Empirical Methods in Natural Language Processing}, 2014,
  pp. 1769--1773.

\bibitem{Hauer2016DecodingAT}
B.~Hauer and G.~Kondrak, ``Decoding anagrammed texts written in an unknown
  language and script,'' \emph{Transactions of the Association for
  Computational Linguistics}, vol.~4, pp. 75--86, 2016.

\end{thebibliography}


\begin{thebibliography}{14}
\expandafter\ifx\csname natexlab\endcsname\relax\def\natexlab#1{#1}\fi

\bibitem[{Berg-Kirkpatrick et~al.(2013)Berg-Kirkpatrick, Durrett, and
  Klein}]{conf/acl/Berg-KirkpatrickDK13}
Taylor Berg-Kirkpatrick, Greg Durrett, and Dan Klein. 2013.
\newblock Unsupervised transcription of historical documents.
\newblock In \emph{ACL (1)}, pages 207--217. The Association for Computer
  Linguistics.

\bibitem[{Breuel(2000)}]{doi:10.1117/12.410858}
Thomas~M. Breuel. 2000.
\newblock Modeling the sample distribution for clustering ocr.

\bibitem[{Ho and Nagy(2000)}]{902858}
Tin~Kam Ho and G.~Nagy. 2000.
\newblock Ocr with no shape training.
\newblock In \emph{Proceedings 15th International Conference on Pattern
  Recognition. ICPR-2000}, volume~4, pages 27--30 vol.4.

\bibitem[{Huang et~al.(2007)Huang, Learned-Miller, and McCallum}]{4378705}
G.~Huang, E.~Learned-Miller, and A.~McCallum. 2007.
\newblock Cryptogram decoding for ocr using numerization strings.
\newblock In \emph{Ninth International Conference on Document Analysis and
  Recognition (ICDAR 2007)}, volume~1, pages 208--212.

\bibitem[{Kamper et~al.(2017)Kamper, Jansen, and
  Goldwater}]{Kamper:2017:SFF:3143821.3144152}
Herman Kamper, Aren Jansen, and Sharon Goldwater. 2017.
\newblock A segmental framework for fully-unsupervised large-vocabulary speech
  recognition.
\newblock \emph{Comput. Speech Lang.}, 46(C):154--174.

\bibitem[{Knight et~al.(2011)Knight, Megyesi, and
  Schaefer}]{Knight:2011:CC:2024236.2024239}
Kevin Knight, Be\'{a}ta Megyesi, and Christiane Schaefer. 2011.
\newblock The copiale cipher.
\newblock In \emph{Proceedings of the 4th Workshop on Building and Using
  Comparable Corpora: Comparable Corpora and the Web}, BUCC '11, pages 2--9,
  Stroudsburg, PA, USA. Association for Computational Linguistics.

\bibitem[{Knight et~al.(2006)Knight, Nair, Rathod, and
  Yamada}]{knight-EtAl:2006:POS}
Kevin Knight, Anish Nair, Nishit Rathod, and Kenji Yamada. 2006.
\newblock Unsupervised analysis for decipherment problems.
\newblock In \emph{Proceedings of the COLING/ACL 2006 Main Conference Poster
  Sessions}, pages 499--506, Sydney, Australia. Association for Computational
  Linguistics.

\bibitem[{Knight and Yamada(1999)}]{Knight99acomputational}
Kevin Knight and Kenji Yamada. 1999.
\newblock A computational approach to deciphering unknown scripts.
\newblock In \emph{in: Proceedings of the ACL Workshop on Unsupervised Learning
  in Natural Language Processing}.

\bibitem[{Koch et~al.(2015)Koch, Zemel, and Salakhutdinov}]{Koch2015SiameseNN}
Gregory Koch, Richard Zemel, and Ruslan Salakhutdinov. 2015.
\newblock Siamese neural networks for one-shot image recognition.

\bibitem[{Nagy(1992)}]{156472}
G.~Nagy. 1992.
\newblock At the frontiers of ocr.
\newblock \emph{Proceedings of the IEEE}, 80(7):1093--1100.

\bibitem[{Nagy et~al.(1987)Nagy, Seth, and
  Einspahr}]{Nagy:1987:DSC:28809.28825}
G.~Nagy, S.~Seth, and K.~Einspahr. 1987.
\newblock Decoding substitution ciphers by means of word matching with
  application to ocr.
\newblock \emph{IEEE Trans. Pattern Anal. Mach. Intell.}, 9(5):710--715.

\bibitem[{Smith(2007)}]{Smith:2007:OTO:1304596.1304846}
R.~Smith. 2007.
\newblock An overview of the tesseract ocr engine.
\newblock In \emph{Proceedings of the Ninth International Conference on
  Document Analysis and Recognition - Volume 02}, ICDAR '07, pages 629--633,
  Washington, DC, USA. IEEE Computer Society.

\bibitem[{Snyder et~al.(2010)Snyder, Barzilay, and
  Knight}]{Snyder:2010:SML:1858681.1858788}
Benjamin Snyder, Regina Barzilay, and Kevin Knight. 2010.
\newblock A statistical model for lost language decipherment.
\newblock In \emph{Proceedings of the 48th Annual Meeting of the Association
  for Computational Linguistics}, ACL '10, pages 1048--1057, Stroudsburg, PA,
  USA. Association for Computational Linguistics.

\bibitem[{Zhang et~al.(2016)Zhang, Chowdhury, Dhulekar, Xia, Knight, Ji, Yener,
  and Zhao}]{nyushu}
Tongtao Zhang, Aritra Chowdhury, Nimit Dhulekar, Jinjing Xia, Kevin Knight,
  Heng Ji, B\"{u}lent Yener, and Liming Zhao. 2016.
\newblock From image to translation: Processing the endangered {Nyushu} script.
\newblock \emph{ACM Transactions on Asian and Low-Resource Language Information
  Processing}, 15(4).

\end{thebibliography}
%

\end{document}